\pdfoutput=1

\documentclass[11pt]{article}

\usepackage{acl}
\usepackage{graphicx}
\usepackage{booktabs}

\usepackage{times}
\usepackage{latexsym}

\usepackage{subfigure}

\usepackage{booktabs, multirow} 
\usepackage{soul}

\usepackage[T1]{fontenc}

\usepackage[utf8]{inputenc}

\usepackage{microtype}

\title{Estimating Semantic Similarity between In-Domain and Out-of-Domain Samples}

\author{Rhitabrat Pokharel \and Ameeta Agrawal\\
  Department of Computer Science\\
  Portland State University \\
  \texttt{\{pokharel,ameeta\}@pdx.edu} \\}

\begin{document}
\maketitle
\begin{abstract}

Prior work typically describes out-of-domain (OOD) or out-of-distribution (OODist) samples as those that originate from dataset(s) or source(s) different from the training set but for the same task. When compared to in-domain (ID) samples, the models have been known to usually perform poorer on OOD samples, although this observation is not consistent. Another thread of research has focused on OOD detection, albeit mostly using supervised approaches. In this work, we first consolidate and present a systematic analysis of multiple definitions of OOD and OODist  as discussed in prior literature. Then, we analyze the  performance of a model under ID and OOD/OODist settings in a principled way. Finally, we seek to identify an unsupervised method for reliably identifying OOD/OODist samples without using a trained model. The results of our extensive evaluation using 12 datasets from 4 different tasks suggest the promising potential of unsupervised metrics in this task.



\end{abstract}

\section{Introduction}
What happens when you train a machine learning model on a dataset and use it to predict a sample whose source is unknown? Would you fully rely on the model's prediction on the test sample? Basically, this situation is encountered in most real-world scenarios where the test sample may differ considerably from the training samples. Recent works show that models perform poorer on the samples that come from a different distribution {\citep{Gokhale2022}}. In many real-world scenarios, such as health and law, false predictions or misclassified results could have significant consequences, and as such  identifying out-of-domain or out-of-distribution data beforehand is critical.


\begin{table*}[!t]
\centering
\begin{tabular}{p{4.3cm} c c p{3.7cm} p{3.9cm}}
\toprule
\textbf{Paper}  & \textbf{Setup}  & \textbf{Term} & \textbf{Metrics} & \textbf{Task}\\
\midrule
\citet{Chrysostomou2022} & A & OOD & - & Sentiment classification\\
\citet{Berre2022}        & A & OOD & Accuracy & MCQ\\
\citet{Lin2022}          & A & OODist & - & Extractive QA\\
\citet{Nejadgholi2022}   & A & OOD & AUC, F1 & Sentiment classification\\
\citet{David2022}        & A & OODist    & Cosine similarity, Confidence score, Probability distribution      & Sentiment classification\\
\citet{Mishra2022}       & A & OODist & NLI diagnostics & NLI\\
\citet{Varshney2022}     & A & OOD   & Accuracy  & NLI, Duplicate detection, Sentiment analysis, MCQ, Commonsense Reasoning\\
\citet{Omar2022}         & A & OODist    & Accuracy, Success rate, Error rate, Diversity, Fairness, IBP tightness, Robustness     & Classification, Paraphrasing, NLI   \\
\citet{Adila2021}        & A & OODist    & Confidence, Variability   & NLI   \\
\citet{singhal2022assessing} & A & OOD & Accuracy & NLI, Phrase identification \\
\citet{agrawal2022rethinking} & A & OOD & Accuracy & Visual QA  \\
\citet{Ehsan2022}        & A, B  & OODist    & Accuracy  & Metaphorical knowledge\\
\citet{chen2023fine} &  A, B & OODist &  Accuracy & Sentiment analysis, Toxicity detection, News Classification, Dialogue Intent
Classification\\
\citet{Mai2022}          & B & OODist & - & Anomaly detection\\
\citet{garg2022leveraging} & B & OOD & Accuracy & {Rating generation, Toxicity classification}\\
\citet{jin2021towards}      & B & OOD   & False Positive Ratio, AUROC, AUPR   & Text Classification\\
\citet{Atwell2022}       & C & OOD & h-discrepancy & Discourse parsing\\
\citet{Gokhale2022}      & C & OOD   & Accuracy, EM  & NLI, QA, Image classification\\
\bottomrule
\end{tabular}
\caption{A survey of recent works using various setups to study OODist or OOD settings. Here, {\bf A} describes the cases where the train set is from one dataset, and the test set from another dataset; {\bf B} describes the scenario where the train and test sets are two subsets of the same dataset; and {\bf C} is a combination of both A and B. The ``Metrics" column represents the metrics, while the ``Task" column lists the tasks studied in these papers. Note that several papers whose setup can be described as {\bf A} use different terms.}
\label{tab:OODist survey}
\end{table*}

Previous works have defined OOD and OODist data in different ways or used them interchangeably. Early works define data that comes from a related but different domain as OOD \citep{Dai2007}, whereas  OODist data has been defined as the data that might have been collected at a different time \citep{Yaniv2019}. In recent studies, \citep{Chrysostomou2022} use the term OOD to describe different datasets for the same task (e.g., SST, IMDb, and Yelp for sentiment classification), whereas \citep{Lin2022} use OODist to describe the datasets that are not in the training set, including those that are subsets of the same dataset (e.g., PDTB 2.0 \citep{Carlson2002}). In this paper, we first present a focused analysis of all the various terminologies used in this context in recent works.

Another thread of research has focused on identifying OOD/OODist samples, mostly through supervised methods \citep{Varshney2022, David2022, Gokhale2022}. However, considering that trained models may not always be available, we take a complementary approach in this work to identify metric(s) that may be able to support OOD detection in an {\em unsupervised} manner. 

The first part of our methodology focuses on establishing to what extent performance (e.g., accuracy) can inform the detection of OOD samples\footnote{As formally distinguishing between the two terms remains beyond the scope of this paper, in this work we use the terms OOD and OODist interchangeably.}. Our results indicate that indeed performance can serve as a reliable metric for estimating OODness, however, this requires a supervised model. To address this limitation, in the second part of this work, we explore several unsupervised metrics for estimating semantic similarity between the training and test samples. We hypothesize that an unsupervised metric which sufficiently correlates with performance, may be considered as a feasible alternative for detecting OOD samples.

The major contributions of this paper are:

\begin{itemize}
\item  an in-depth exploration of the usage of the terms OOD and OODist in recent works; 

 \item  a systematic assessment of the effectiveness of performance in estimating OODness, and an investigation of unsupervised approaches for identifying OODness;

\item    an extensive evaluation across four different tasks using a total of twelve datasets; we will also make our code available for facilitating reproducibility.

\end{itemize}

\begin{table*}[!t]
\centering
\setlength{\tabcolsep}{20pt}
\begin{tabular}{l l l}
\toprule
\textbf{Task}  & \textbf{Datasets}    & \textbf{train/ val/ test} \\
\midrule
Sentiment  & IMDb, SST2, Yelp  & 3310/ 428/ 909\\
MCQ  & SCIQ, CS, QASC    & 8134/ 926/ 920\\
Extractive QA       & SQUAD, News, Trivia  & 61688/ -/ 4212\\
NLI       & MNLI, WNLI, QNLI  & 635/ 71/ 146\\
\bottomrule
\end{tabular}
\caption{Task and dataset details}
\label{tab:dataset_details}
\end{table*}


\section{Related Work}


Prior research has often used the terms OOD and OODist interchangeably. In some works, dataset $X$ is described to be OODist to dataset $Y$ if they are different datasets, but support the same task \citep{Lin2022, Ehsan2022, David2022, Mishra2022, Omar2022, Adila2021}, while in other works, the term OOD is used to describe the similar setting \cite{Chrysostomou2022, Berre2022, Nejadgholi2022, Varshney2022}. Beyond that, while some consider different subsets of the same dataset to be OODist \cite{Mai2022, garg2022leveraging, jin2021towards}, others refer to these as OOD to describe distributionally different datasets \cite{Atwell2022, Gokhale2022}.






When it comes to detecting OOD or OODist samples, using the model's accuracy \citep{Berre2022, Ehsan2022, Gokhale2022, Omar2022}, input features, hidden features representations, and output probability distribution of the network layers \cite{David2022}, or AUC and F1 score \cite{Nejadgholi2022} have been well-studied. Table \ref{tab:OODist survey} presents a brief summary of some recent works. 

\section{Method}

\subsection{Problem Definition}
Given two datasets, $\mathcal{X}=\{x_1, ..., x_m\}$ and $\mathcal{Y}=\{y_1, ..., y_m\}$, the goal is to assess the correlation between the performance of the two datasets under ID/OOD settings and their (semantic) similarity. The performance is measured by training a model on one of the datasets, say, $\mathcal{X}_{train}$ and testing it on the test set $\mathcal{X}_{test}$ which represents the ID setting, and $\mathcal{Y}_{test}$ representing the OOD setting. The ID similarity is computed by averaging the similarity between the instances of $\mathcal{X}_{train}$ and $\mathcal{X}_{test}$, while OOD similarity is measured between $\mathcal{X}_{train}$ and $\mathcal{Y}_{test}$.




\subsection{Datasets} 
{We study four different tasks using a total of 12 datasets (3 datasets for per task).} We include the most common tasks that have been used in prior work. 

\noindent \textbf{\em (i) Sentiment Analysis}: given a text, classify its sentiment as negative or positive. 

\noindent \textbf{\em  (ii) Multiple Choice Question Answering (MCQ)}: given a question and a context, select the correct answer from a pool of possible answers. 

\noindent \textbf{\em  (iii) Extractive Question Answering (QA)}: given a question and a context, find the answer to the question from the context. 

\noindent \textbf{\em (iv) Natural Language Inference (NLI)}: given a premise and a hypothesis, {determine whether the hypothesis contradicts, entails, or is neutral with respect to the premise.}

{Table \ref{tab:dataset_details} presents the details of the datasets and the tasks. For sentiment classification, we use IMDb \citep{Maas2011}, SST2 \citep{Socher2013}, and Yelp \citep{Zhang2015} datasets. We experiment with SCIQ \cite{Johannes2017}, CommonsenseQA (CS) \citep{Talmor2019}, and QASC \citep{Tushar2020} for the MCQ task. For the Extractive QA task, SQUAD, News, and Trivia \citep{Fisch2019} datasets are selected from the MRQA dataset (note that since these datasets do not have a separate test set, we use the validation data as the test set). {The NLI datasets include MNLI, QNLI, and WNLI from the GLUE benchmark \cite{wang2018glue}}. All the other datasets were accessed from  the HuggingFace repository\footnote{\url{https://huggingface.co/datasets/}}.

\medskip
\noindent \textbf{Data preparation}: Prior work has largely overlooked the effect of an important aspect -- dataset size -- in such studies. As such, we control the dataset size as a variable in our study by maintaining the size of all train, validation (when available), and test splits for all three datasets per task by downsampling them to match the size of the {\bf smallest dataset in each set}. For instance, all the splits of all three sentiment analysis datasets are downsampled to be of equal size. Additionally, we balance the number of instances for each class when possible (e.g., in the sentiment datasets). 


\subsection{Metrics} \label{section:metrics}

We use three categories of metrics, one for measuring the performance of the model, another for estimating the similarity between the two datasets, and the third for computing the correlation between performance and similarity.

\medskip
\noindent \textbf{Performance Metrics}. We report accuracy for the classification tasks, i.e., sentiment analysis, MCQ, and NLI tasks, and F1 score for extractive Question Answering task to measure the correctness of model predictions.


\begin{table}[!t]
\centering
\setlength{\tabcolsep}{6pt}
\begin{tabular}{lrc}
\toprule
\textbf{Trained on} &\textbf{Tested on} &\textbf{Performance} \\
\midrule
\multirow{3}{*}{\textcolor{blue}{IMDb-train}} &IMDb-test &0.90 \\
 &Yelp-test &0.87 \\
 &SST2-test &0.17 \\
 \cmidrule{2-3}
\multirow{3}{*}{\textcolor{blue}{SST2-train}} &SST2-test &0.89 \\
 &IMDb-test &0.21 \\
 &Yelp-test &0.16 \\
 \cmidrule{2-3}
\multirow{3}{*}{\textcolor{blue}{Yelp-train}} &Yelp-test &0.93 \\
 &IMDb-test &0.86 \\
 &SST2-test &0.19 \\
\midrule
\multirow{3}{*}{\textcolor{blue}{SCIQ-train}} &SCIQ-test &0.64 \\
 &QASC-test &0.18 \\
 &CS-test &0.34 \\
 \cmidrule{2-3}
\multirow{3}{*}{CS-train} &CS-test &0.49 \\
 &SCIQ-test &0.58 \\
&QASC-test &0.84 \\
\cmidrule{2-3}
\multirow{3}{*}{\textcolor{blue}{QASC-train}} &QASC-test &0.92 \\
&SCIQ-test &0.51 \\
 &CS-test &0.48 \\
\midrule
\multirow{3}{*}{\textcolor{blue}{SQUAD-train}} &SQUAD-test &0.86 \\
 &News-test &0.51 \\
&Trivia-test &0.55 \\
\cmidrule{2-3}
\multirow{3}{*}{News-train} &News-test &0.66 \\
&SQUAD-test &0.77 \\
 &Trivia-test &0.56 \\
 \cmidrule{2-3}
\multirow{3}{*}{\textcolor{blue}{Trivia-train}} &Trivia-test &0.66 \\
 &SQUAD-test &0.52 \\
 &News-test &0.31 \\
\midrule
\multirow{3}{*}{\textcolor{blue}{MNLI-train}} &MNLI-test &0.57 \\
 &WNLI-test &0.56 \\
 &QNLI-test &0.54 \\
 \cmidrule{2-3}
\multirow{3}{*}{WNLI-train} &WNLI-test &0.42 \\
 &MNLI-test &0.26 \\
 &QNLI-test &0.47 \\
 \cmidrule{2-3}
\multirow{3}{*}{\textcolor{blue}{QNLI-train}} &QNLI-test &0.83 \\
 &MNLI-test &0.43 \\
 &WNLI-test &0.56 \\
\bottomrule
\end{tabular}
\caption{Performance results under different ID/OOD settings. Instances where ID performance is better than OOD performance are indicated in \textcolor{blue}{blue}.}
\label{tab:acc_results}
\end{table}

\medskip
\noindent \textbf{Similarity Metrics}. To estimate the closeness among the ID and OOD datasets, we use metrics related to semantic similarity (higher value means the samples are from nearby distributions) and semantic distance (higher value indicates less similarity). These include: \textbf{\em (i) Cosine Similarity}: measures the distance between the samples from two sources\footnote{We estimate this using \texttt{word2vec} embeddings.}. \textbf{\em (ii) Mauve Score}: measures the similarity between two texts\footnote{We use the default embeddings (GPT-2) \url{https://pypi.org/project/mauve-text/}.} \citep{Pillutla2021}. \textbf{\em (iii) Wasserstein Distance (Wstn)}: measures the distance between the two distributions and if the distributions overlap enough, then they are close to each other\footnote{We use the universal sentence encoder for estimating this.} \citep{Weng2019}. \textbf{\em (iv) Jensen Shannon Distance (JSD)}: quantifies the similarity between two probability distributions, where the smaller the value, the closer the distributions\footnote{We used \texttt{word2vec} embeddings.} \citep{Manning1999}. 

\medskip
\noindent \textbf{Correlation Metrics}. Lastly, we use two commonly used correlation metrics -- Kendall Tau and Pearson\footnote{\url{https://pandas.pydata.org/docs/reference/api/pandas.DataFrame.corr.html}} (we also experimented with Spearman which gave similar results), with the goal of understanding the  relationship between performance and similarity of datasets under ID/OOD settings.


\subsection{Measuring Performance and Similarity}
For measuring the performance, we fine-tune a BERT \texttt{base} uncased model for 2 epochs on each $\mathcal{X}_{train}$ and test it on $\mathcal{X}_{test}$ (ID) and $\mathcal{Y}_{test}$ (OOD). For estimating the similarity between the ID and OOD datasets, we randomly sample two sets of 20 instances,  $\mathcal{X}_{train20}$ and $\mathcal{Y}_{test20}$, and estimate pairwise similarity between all of these samples, obtaining a total of 400 similarity scores which are then averaged to compute the similarity. 


\section{Results and Discussion}



\textbf{\em Performance analysis}: {Table~\ref{tab:acc_results} presents the results of the performance experiments, where we observe that the model performance under ID settings is generally better than under OOD settings, except for three exceptions, suggesting that performance can indeed serve as a reasonably dependable metric for detecting OOD. However, this requires a supervised model, which motivates us to explore unsupervised approaches for estimating OODness. It is worth noting that while \citet{garg2022leveraging} found that OOD accuracy is less than the ID accuracy, this observation does not always hold true according to our analysis.




\medskip

\noindent \textbf{\em Correlation between performance and similarity}: Figure \ref{fig:acc_vs_metrics}  presents the heatmap visualizing the correlation (Kendall and Pearson) between performance and similarity metrics, across all 12 datasets for the four tasks (the full set of results is included in Appendix~\ref{sec:appendix}). In looking at the results, we observe that according to Kendall Tau correlation analysis, Wasserstein distance (Wstn) shows the most consistent correlation (in 10 out of 12 cases), whereas according to Pearson correlation, both Wasserstein and Cosine are acceptable metrics (in 9 out of 12 cases). In all the scenarios, however, JSD is clearly the least correlated metric. This suggests the potential of unsupervised approaches in estimating OOD samples.

\begin{figure}[t]
    \centering
    \subfigure[]{\includegraphics[width=.49\textwidth]{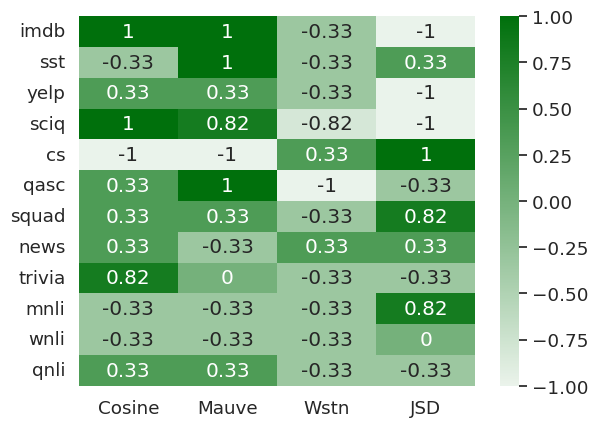}}
    \subfigure[]{\includegraphics[width=.49\textwidth]{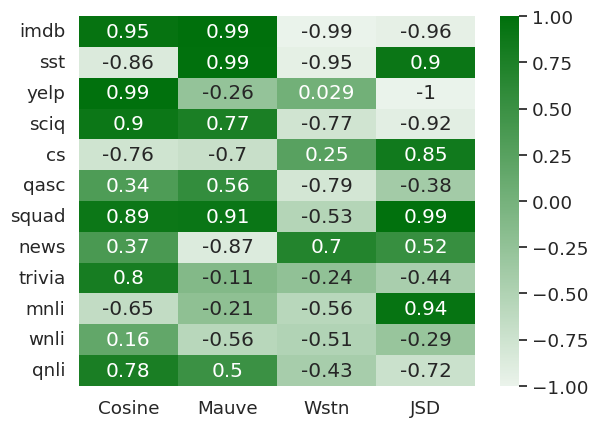}}
    \caption{(a) Kendall and (b) Pearson  correlation between performance and dataset similarity, evaluated over 12 datasets with each serving as an ID dataset once. For Cosine and Mauve, darker shades are desirable, whereas for Wstn and JSD, lighter shades indicate better correlation.}
    \label{fig:acc_vs_metrics}
    
\end{figure}

\section{Conclusion}
In this work, we aim to identify unsupervised approaches for identifying OOD samples. We conducted an in-depth analysis of different unsupervised similarity metrics and estimated their correlation with performance of a model under ID/OOD settings. Our findings indicate that Wasserstein distance presents a promising metric for determining OOD samples. The natural question of how to determine the appropriate threshold, however, remains to be explored in future work. Another direction worth exploring is to verify the robustness of these similarity metrics when estimated using different embeddings.

\section*{Limitations}
While our analysis suggests some promising results, we acknowledge some limitations of this work such as:
\begin{itemize}
    \item on some datasets, the ID performance was observed to be less than the OOD performance, and further investigation is needed to study this observation in detail and bring additional insights.
    \item all the analysis in this study focuses on datasets in English language, and it will be interesting to investigate whether our findings will generalize to other languages.
\end{itemize}



\section*{Acknowledgements}
We are thankful to the anonymous reviewers for their valuable feedback.

\bibliography{acl_latex}

\begin{thebibliography}{32}
\expandafter\ifx\csname natexlab\endcsname\relax\def\natexlab#1{#1}\fi

\bibitem[{Adila and Kang(2022)}]{Adila2021}
Dyah Adila and Dongyeop Kang. 2022.
\newblock \href {https://proceedings.mlr.press/v163/adila22a.html}
  {Understanding out-of-distribution: A perspective of data dynamics}.
\newblock In \emph{Proceedings on "I (Still) Can't Believe It's Not Better!" at
  NeurIPS 2021 Workshops}, volume 163 of \emph{Proceedings of Machine Learning
  Research}, pages 1--8. PMLR.

\bibitem[{Aghazadeh et~al.(2022)Aghazadeh, Fayyaz, and
  Yaghoobzadeh}]{Ehsan2022}
Ehsan Aghazadeh, Mohsen Fayyaz, and Yadollah Yaghoobzadeh. 2022.
\newblock \href {https://doi.org/10.48550/ARXIV.2203.14139} {Metaphors in
  pre-trained language models: Probing and generalization across datasets and
  languages}.

\bibitem[{Agrawal et~al.(2022)Agrawal, Kaji{\'c}, Bugliarello, Davoodi,
  Gergely, Blunsom, and Nematzadeh}]{agrawal2022rethinking}
Aishwarya Agrawal, Ivana Kaji{\'c}, Emanuele Bugliarello, Elnaz Davoodi, Anita
  Gergely, Phil Blunsom, and Aida Nematzadeh. 2022.
\newblock Rethinking evaluation practices in visual question answering: A case
  study on out-of-distribution generalization.
\newblock \emph{arXiv preprint arXiv:2205.12191}.

\bibitem[{Atwell et~al.(2022)Atwell, Sicilia, Hwang, and Alikhani}]{Atwell2022}
Katherine Atwell, Anthony Sicilia, Seong~Jae Hwang, and Malihe Alikhani. 2022.
\newblock \href {https://doi.org/10.18653/v1/2022.findings-acl.68} {The change
  that matters in discourse parsing: Estimating the impact of domain shift on
  parser error}.
\newblock In \emph{Findings of the Association for Computational Linguistics:
  ACL 2022}, pages 824--845, Dublin, Ireland. Association for Computational
  Linguistics.

\bibitem[{Carlson et~al.(2002)Carlson, Okurowski, and Marcu}]{Carlson2002}
Lynn Carlson, Mary~Ellen Okurowski, and Daniel Marcu. 2002.
\newblock \emph{RST discourse treebank}.
\newblock Linguistic Data Consortium, University of Pennsylvania.

\bibitem[{Chen et~al.(2023)Chen, Yang, Bi, and Sun}]{chen2023fine}
Sishuo Chen, Wenkai Yang, Xiaohan Bi, and Xu~Sun. 2023.
\newblock Fine-tuning deteriorates general textual out-of-distribution
  detection by distorting task-agnostic features.
\newblock \emph{arXiv preprint arXiv:2301.12715}.

\bibitem[{Chiang and Lee(2022)}]{David2022}
David Cheng-Han Chiang and Hung-Yi Lee. 2022.
\newblock \href {https://doi.org/10.48550/arXiv.2204.04458} {Understanding,
  detecting, and separating out-of-distribution samples and adversarial samples
  in text classification}.
\newblock \emph{CoRR}, abs/2204.04458.

\bibitem[{Chrysostomou and Aletras(2022)}]{Chrysostomou2022}
George Chrysostomou and Nikolaos Aletras. 2022.
\newblock \href {https://doi.org/10.48550/ARXIV.2203.00056} {An empirical study
  on explanations in out-of-domain settings}.

\bibitem[{Dai et~al.(2007)Dai, Xue, Yang, and Yu}]{Dai2007}
Wenyuan Dai, Gui-Rong Xue, Qiang Yang, and Yong Yu. 2007.
\newblock Co-clustering based classification for out-of-domain documents.
\newblock In \emph{Proceedings of the 13th ACM SIGKDD international conference
  on Knowledge discovery and data mining}, pages 210--219.

\bibitem[{Fisch et~al.(2019)Fisch, Talmor, Jia, Seo, Choi, and
  Chen}]{Fisch2019}
Adam Fisch, Alon Talmor, Robin Jia, Minjoon Seo, Eunsol Choi, and Danqi Chen.
  2019.
\newblock {MRQA} 2019 shared task: Evaluating generalization in reading
  comprehension.
\newblock In \emph{Proceedings of 2nd Machine Reading for Reading Comprehension
  (MRQA) Workshop at EMNLP}.

\bibitem[{Garg et~al.(2022)Garg, Balakrishnan, Lipton, Neyshabur, and
  Sedghi}]{garg2022leveraging}
Saurabh Garg, Sivaraman Balakrishnan, Zachary~C Lipton, Behnam Neyshabur, and
  Hanie Sedghi. 2022.
\newblock Leveraging unlabeled data to predict out-of-distribution performance.
\newblock \emph{arXiv preprint arXiv:2201.04234}.

\bibitem[{Gokhale et~al.(2022)Gokhale, Mishra, Luo, Sachdeva, and
  Baral}]{Gokhale2022}
Tejas Gokhale, Swaroop Mishra, Man Luo, Bhavdeep Sachdeva, and Chitta Baral.
  2022.
\newblock \href {https://doi.org/10.18653/v1/2022.findings-acl.213}
  {\textit{Generalized but not Robust?} comparing the effects of data
  modification methods on out-of-domain generalization and adversarial
  robustness}.
\newblock In \emph{Findings of the Association for Computational Linguistics:
  ACL 2022}, pages 2705--2718, Dublin, Ireland. Association for Computational
  Linguistics.

\bibitem[{Jin et~al.(2021)Jin, Gao, Kim, Liu, and
  Hakkani-Tür}]{jin2021towards}
Di~Jin, Shuyang Gao, Seokhwan Kim, Yang Liu, and Dilek Hakkani-Tür. 2021.
\newblock \href
  {https://www.amazon.science/publications/towards-textual-out-of-domain-detection-without-in-domain-labels}
  {Towards textual out-of-domain detection without in-domain labels}.
\newblock \emph{IEEE/ACM Transactions on Audio, Speech, and Language
  Processing}.

\bibitem[{Khot et~al.(2020)Khot, Clark, Guerquin, Jansen, and
  Sabharwal}]{Tushar2020}
Tushar Khot, Peter Clark, Michal Guerquin, Peter Jansen, and Ashish Sabharwal.
  2020.
\newblock Qasc: A dataset for question answering via sentence composition.
\newblock \emph{arXiv:1910.11473v2}.

\bibitem[{Le~Berre et~al.(2022)Le~Berre, Cerisara, Langlais, and
  Lapalme}]{Berre2022}
Guillaume Le~Berre, Christophe Cerisara, Philippe Langlais, and Guy Lapalme.
  2022.
\newblock \href {https://doi.org/10.18653/v1/2022.acl-short.83} {Unsupervised
  multiple-choice question generation for out-of-domain {Q}{\&}{A}
  fine-tuning}.
\newblock In \emph{Proceedings of the 60th Annual Meeting of the Association
  for Computational Linguistics (Volume 2: Short Papers)}, pages 732--738,
  Dublin, Ireland. Association for Computational Linguistics.

\bibitem[{Lin et~al.(2022)Lin, Wang, Lin, Jia, Xiao, Ren, and Yih}]{Lin2022}
Bill~Yuchen Lin, Sida Wang, Xi~Lin, Robin Jia, Lin Xiao, Xiang Ren, and Scott
  Yih. 2022.
\newblock \href {https://doi.org/10.18653/v1/2022.acl-long.223} {On continual
  model refinement in out-of-distribution data streams}.
\newblock In \emph{Proceedings of the 60th Annual Meeting of the Association
  for Computational Linguistics (Volume 1: Long Papers)}, pages 3128--3139,
  Dublin, Ireland. Association for Computational Linguistics.

\bibitem[{Maas et~al.(2011)Maas, Daly, Pham, Huang, Ng, and Potts}]{Maas2011}
Andrew~L. Maas, Raymond~E. Daly, Peter~T. Pham, Dan Huang, Andrew~Y. Ng, and
  Christopher Potts. 2011.
\newblock \href {http://www.aclweb.org/anthology/P11-1015} {Learning word
  vectors for sentiment analysis}.
\newblock In \emph{Proceedings of the 49th Annual Meeting of the Association
  for Computational Linguistics: Human Language Technologies}, pages 142--150,
  Portland, Oregon, USA. Association for Computational Linguistics.

\bibitem[{Mai et~al.(2022)Mai, Davies, and Griffin}]{Mai2022}
Kimberly~T. Mai, Toby Davies, and Lewis~D. Griffin. 2022.
\newblock \href {http://arxiv.org/abs/2204.05695} {Self-supervised losses for
  one-class textual anomaly detection}.

\bibitem[{Manning and Schutze(1999)}]{Manning1999}
Christopher Manning and Hinrich Schutze. 1999.
\newblock \emph{Foundations of statistical natural language processing}.
\newblock MIT press.

\bibitem[{Mishra and Arunkumar(2022)}]{Mishra2022}
Swaroop Mishra and Anjana Arunkumar. 2022.
\newblock \href {https://doi.org/10.48550/ARXIV.2203.06404} {A proposal to
  study "is high quality data all we need?"}.

\bibitem[{Nejadgholi et~al.(2022)Nejadgholi, Fraser, and
  Kiritchenko}]{Nejadgholi2022}
Isar Nejadgholi, Kathleen Fraser, and Svetlana Kiritchenko. 2022.
\newblock \href {https://doi.org/10.18653/v1/2022.acl-long.378} {Improving
  generalizability in implicitly abusive language detection with concept
  activation vectors}.
\newblock In \emph{Proceedings of the 60th Annual Meeting of the Association
  for Computational Linguistics (Volume 1: Long Papers)}, pages 5517--5529,
  Dublin, Ireland. Association for Computational Linguistics.

\bibitem[{{Omar} et~al.(2022){Omar}, {Choi}, {Nyang}, and
  {Mohaisen}}]{Omar2022}
Marwan {Omar}, Soohyeon {Choi}, DaeHun {Nyang}, and David {Mohaisen}. 2022.
\newblock \href {http://arxiv.org/abs/2201.00768} {{Robust Natural Language
  Processing: Recent Advances, Challenges, and Future Directions}}.
\newblock \emph{arXiv e-prints}, page arXiv:2201.00768.

\bibitem[{Ovadia et~al.(2019)Ovadia, Fertig, Ren, Nado, Sculley, Nowozin,
  Dillon, Lakshminarayanan, and Snoek}]{Yaniv2019}
Yaniv Ovadia, Emily Fertig, Jie Ren, Zachary Nado, David Sculley, Sebastian
  Nowozin, Joshua Dillon, Balaji Lakshminarayanan, and Jasper Snoek. 2019.
\newblock Can you trust your model's uncertainty? evaluating predictive
  uncertainty under dataset shift.
\newblock \emph{Advances in neural information processing systems}, 32.

\bibitem[{Pillutla et~al.(2021)Pillutla, Swayamdipta, Zellers, Thickstun,
  Welleck, Choi, and Harchaoui}]{Pillutla2021}
Krishna Pillutla, Swabha Swayamdipta, Rowan Zellers, John Thickstun, Sean
  Welleck, Yejin Choi, and Zaid Harchaoui. 2021.
\newblock \href {https://openreview.net/forum?id=Tqx7nJp7PR} {{MAUVE}:
  Measuring the gap between neural text and human text using divergence
  frontiers}.
\newblock In \emph{Advances in Neural Information Processing Systems}.

\bibitem[{Singhal et~al.(2022)Singhal, Forristal, Ye, and
  Durrett}]{singhal2022assessing}
Prasann Singhal, Jarad Forristal, Xi~Ye, and Greg Durrett. 2022.
\newblock Assessing out-of-domain language model performance from few examples.
\newblock \emph{arXiv preprint arXiv:2210.06725}.

\bibitem[{Socher et~al.(2013)Socher, Perelygin, Wu, Chuang, Manning, Ng, and
  Potts}]{Socher2013}
Richard Socher, Alex Perelygin, Jean Wu, Jason Chuang, Christopher~D. Manning,
  Andrew Ng, and Christopher Potts. 2013.
\newblock \href {https://www.aclweb.org/anthology/D13-1170} {Recursive deep
  models for semantic compositionality over a sentiment treebank}.
\newblock In \emph{Proceedings of the 2013 Conference on Empirical Methods in
  Natural Language Processing}, pages 1631--1642, Seattle, Washington, USA.
  Association for Computational Linguistics.

\bibitem[{Talmor et~al.(2019)Talmor, Herzig, Lourie, and Berant}]{Talmor2019}
Alon Talmor, Jonathan Herzig, Nicholas Lourie, and Jonathan Berant. 2019.
\newblock \href {https://doi.org/10.18653/v1/N19-1421} {{C}ommonsense{QA}: A
  question answering challenge targeting commonsense knowledge}.
\newblock In \emph{Proceedings of the 2019 Conference of the North {A}merican
  Chapter of the Association for Computational Linguistics: Human Language
  Technologies, Volume 1 (Long and Short Papers)}, pages 4149--4158,
  Minneapolis, Minnesota. Association for Computational Linguistics.

\bibitem[{Varshney et~al.(2022)Varshney, Mishra, and Baral}]{Varshney2022}
Neeraj Varshney, Swaroop Mishra, and Chitta Baral. 2022.
\newblock \href {https://doi.org/10.18653/v1/2022.acl-long.240} {{ILDAE}:
  Instance-level difficulty analysis of evaluation data}.
\newblock In \emph{Proceedings of the 60th Annual Meeting of the Association
  for Computational Linguistics (Volume 1: Long Papers)}, pages 3412--3425,
  Dublin, Ireland. Association for Computational Linguistics.

\bibitem[{Wang et~al.(2018)Wang, Singh, Michael, Hill, Levy, and
  Bowman}]{wang2018glue}
Alex Wang, Amanpreet Singh, Julian Michael, Felix Hill, Omer Levy, and Samuel~R
  Bowman. 2018.
\newblock Glue: A multi-task benchmark and analysis platform for natural
  language understanding.
\newblock \emph{arXiv preprint arXiv:1804.07461}.

\bibitem[{Welbl et~al.(2017)Welbl, Liu, and Gardner}]{Johannes2017}
Johannes Welbl, Nelson~F Liu, and Matt Gardner. 2017.
\newblock Crowdsourcing multiple choice science questions.
\newblock \emph{arXiv preprint arXiv:1707.06209}.

\bibitem[{Weng(2019)}]{Weng2019}
Lilian Weng. 2019.
\newblock From gan to wgan.
\newblock \emph{ArXiv}, abs/1904.08994.

\bibitem[{Zhang et~al.(2015)Zhang, Zhao, and LeCun}]{Zhang2015}
Xiang Zhang, Junbo Zhao, and Yann LeCun. 2015.
\newblock \href {http://arxiv.org/abs/1509.01626} {Character-level
  {{Convolutional Networks}} for {{Text Classification}}}.
\newblock \emph{arXiv:1509.01626 [cs]}.

\end{thebibliography}
\bibliographystyle{acl_natbib}

\appendix
\section{Experimental Results} \label{sec:appendix}

\begin{table*}[!htp]
\setlength{\tabcolsep}{14pt}
\centering
\begin{tabular}{l l p{1.3cm}r r p{1.2cm}r r r r r}
\toprule
\textbf{Trained} &\textbf{Tested} &\textbf{Model Accuracy} &\textbf{Cosine} &\textbf{Mauve} &\textbf{Wstn} &\textbf{JSD} \\
\midrule
IMDb &IMDb &0.90 &0.92 &1 &0.004 &0.21 \\
IMDb &Yelp &0.87 &0.87 &0.91 &0.0039 &0.26 \\
IMDb &SST2 &0.17 &0.78 &0.42 &0.0052 &0.36 \\
& & & & & & \\
SST2 &SST2 &0.89 &0.66 &0.99 &0.0032 &0.46 \\
SST2 &IMDb &0.21 &0.77 &0.22 &0.0051 &0.38 \\
SST2 &Yelp &0.16 &0.72 &0.004 &0.0046 &0.41 \\
& & & & & & \\
Yelp &Yelp &0.93 &0.86 &0.98 &0.0036 &0.26 \\
Yelp &IMDb &0.86 &0.87 &0.76 &0.0041 &0.27 \\
Yelp &SST2 &0.19 &0.73 &0.94 &0.0038 &0.4 \\
\midrule
SCIQ &SCIQ &0.64 &0.82 &1 &0.004 &0.33 \\
SCIQ &QASC &0.18 &0.66 &0.01 &0.008 &0.46 \\
SCIQ &CS &0.34 &0.78 &1 &0.004 &0.37 \\
& & & & & & \\
CS &CS &0.49 &0.71 &0.94 &0.003 &0.45 \\
CS &SCIQ &0.58 &0.62 &0.01 &0.007 &0.48 \\
CS &QASC &0.84 &0.61 &0.004 &0.005 &0.49 \\
& & & & & & \\
QASC &QASC &0.92 &0.75 &1 &0.003 &0.4 \\
QASC &SCIQ &0.51 &0.78 &0.99 &0.004 &0.38 \\
QASC &CS &0.48 &0.66 &0.004 &0.006 &0.48 \\
\midrule
SQUAD &SQUAD &0.86 &0.84 &0.99 &0.0037 &0.34 \\
SQUAD &NEWS &0.51 &0.82 &0.32 &0.0041 &0.33 \\
SQUAD &TRIVIA &0.55 &0.81 &0.04 &0.0059 &0.33 \\
& & & & & & \\
NEWS &NEWS &0.66 &0.89 &0.91 &0.0036 &0.23 \\
NEWS &SQUAD &0.77 &0.86 &0.11 &0.0046 &0.31 \\
NEWS &TRIVIA &0.56 &0.84 &0.89 &0.0039 &0.27 \\
& & & & & & \\
TRIVIA &TRIVIA &0.66 &0.88 &0.99 &0.0031 &0.23 \\
TRIVIA &SQUAD &0.52 &0.82 &0.04 &0.0062 &0.34 \\
TRIVIA &NEWS &0.31 &0.82 &0.99 &0.0042 &0.29 \\
\midrule
MNLI &MNLI &0.57 &0.72 &0.97 &0.0035 &0.43 \\
MNLI &WNLI &0.56 &0.71 &0.27 &0.0032 &0.43 \\
MNLI &QNLI &0.54 &0.73 &0.99 &0.0037 &0.42 \\
& & & & & & \\
WNLI &WNLI &0.42 &0.74 &0.79 &0.0032 &0.41 \\
WNLI &MNLI &0.26 &0.68 &0.66 &0.0036 &0.46 \\
WNLI &QNLI &0.47 &0.67 &0.004 &0.0035 &0.46 \\
& & & & & & \\
QNLI &QNLI &0.83 &0.75 &0.97 &0.0036 &0.41 \\
QNLI &MNLI &0.43 &0.64 &0.66 &0.0039 &0.45 \\
QNLI &WNLI &0.56 &0.58 &0.01 &0.0034 &0.48 \\
\bottomrule
\end{tabular}
\caption{The results for the sentiment, MCQ, extractive QA, and NLI datasets.}
\label{tab:actual_results}
\end{table*}

\end{document}